\pdfoutput=1

\documentclass[11pt]{article}

\usepackage[]{acl}
\usepackage{graphics}
\usepackage{times}
\usepackage{latexsym}
\usepackage{multirow}
\usepackage{float}
\usepackage{lipsum}
\usepackage{dblfloatfix}

\usepackage[T1]{fontenc}

\usepackage[utf8]{inputenc}

\usepackage{microtype}

\title{TabSQLify: Enhancing Reasoning Capabilities of LLMs Through Table Decomposition}
\author{Md Mahadi Hasan Nahid \\ University of Alberta \\mnahid@ualberta.ca
        \And  
        Davood Rafiei \\ University of Alberta \\ drafiei@ualberta.ca}

\begin{document}

\maketitle
\vspace{-3mm}
\begin{abstract}
Table reasoning is a challenging task that requires understanding both natural language questions and structured tabular data. Large language models (LLMs) have shown impressive capabilities in natural language understanding and generation, but they often struggle with large tables due to their limited input length. In this paper, we propose \textbf{TabSQLify}, a novel method that leverages text-to-SQL generation to decompose tables into smaller and relevant sub-tables, containing only essential information for answering questions or verifying statements, before performing the reasoning task. 
In our comprehensive evaluation on four challenging datasets,  
our approach demonstrates comparable or superior performance compared to prevailing methods reliant on full tables as input. 
Moreover, our method can reduce the input context length significantly, making it more scalable and efficient for large-scale table reasoning applications. Our method performs remarkably well on the WikiTQ benchmark, achieving an accuracy of 64.7\%. Additionally, on the TabFact benchmark, it achieves a high accuracy of 79.5\%. These results surpass other LLM-based baseline models on gpt-3.5-turbo (chatgpt). TabSQLify can reduce the table size significantly alleviating the computational load on LLMs when handling large tables without compromising performance.

\end{abstract}

\section{Introduction}
\begin{figure}[t]
    \centering
    \resizebox{\columnwidth}{!}{%
    \includegraphics{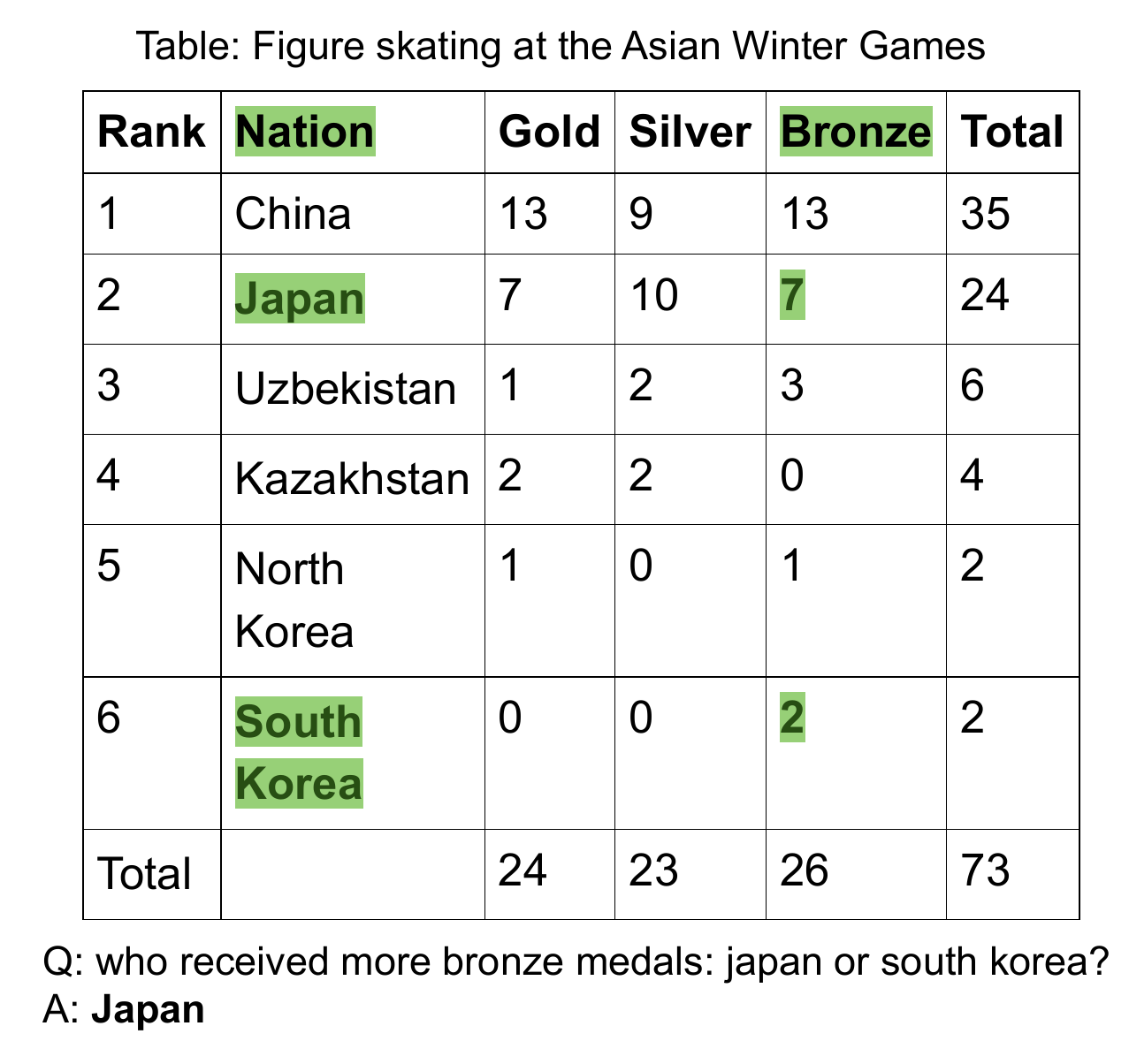}}
    \caption{An example of table-based question answering.}
    \label{fig:example}
\end{figure}

Tables serve as the most prevalent forms of structured information across diverse domains, ranging from databases and spreadsheets to open data repositories, web pages and document collections. 
Developing natural language interfaces for tabular data poses a significant challenge, primarily in terms of effectively interpreting the semantics of table cells and understanding the relationships between cell values in response to a user query. This challenge is accentuated when tables are enveloped in text, such as titles, captions, and contextual text within a document. In these instances, the scope of reasoning expands beyond the confines of table cells to incorporate the surrounding natural language text.
This reasoning is essential for many downstream tasks such as table-based fact verification and table-based question answering (TableQA). As depicted in Figure \ref{fig:example}, table-based reasoning is intricate, demanding sophisticated textual, numerical, and logical reasoning across both unstructured text and (semi-)structured tables. 


\begin{figure*}[t]
    \centering
    \resizebox{\textwidth}{!}{
    \includegraphics{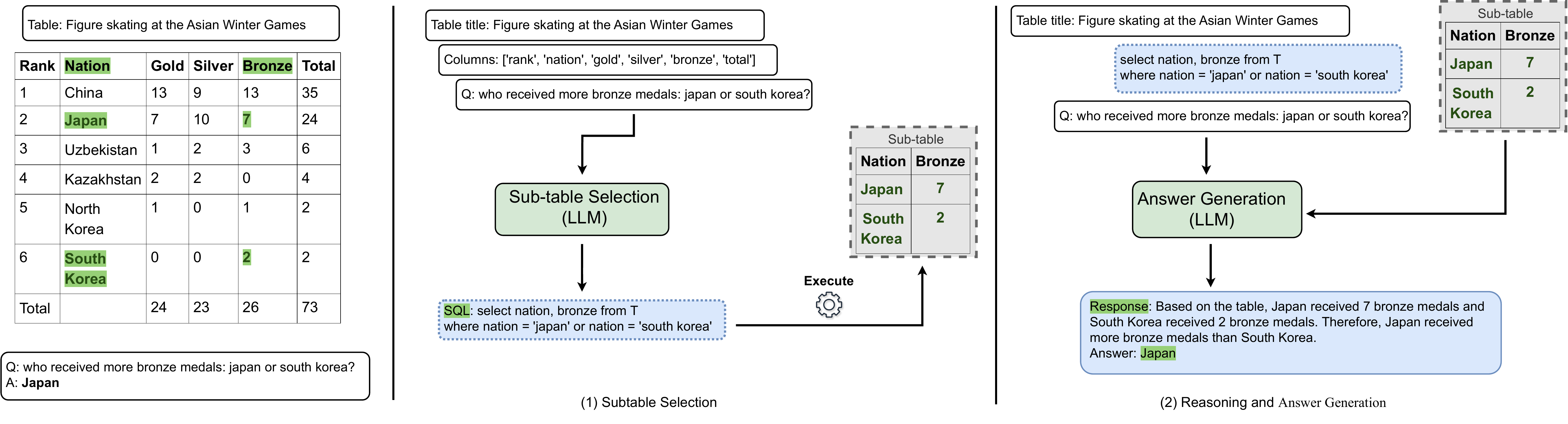}
    }
    \caption{Overview of \textbf{TabSQLify}, consisting of two steps: \textbf{(1)} generating SQL queries from natural language questions or statements and executing the SQL queries on the original tables to obtain sub-tables containing only essential information, and \textbf{(2)} using LLMs with the sub-table and the question or claim to generate the answer.}
    \label{fig:TabSQLify}
\end{figure*}

Recent studies highlight the impressive capability of LLMs in reasoning over both text and tabular data. However, these works typically utilize the full table as context for reasoning~\citep{chen-2023-large}, limiting their ability to large tables. In particular, LLMs operate under a maximum token limit, and when processing a large table, there is a risk of potential truncation of the input or hallucination in the output \citep{chen-2023-large, 10.1145/3571730}. 
This limitation poses difficulties in handling large tables, making it impractical to encompass the entire table within the maximum token boundary of a prompt. \citet{chen-2023-large} highlights that LLMs struggle to generalize when confronted with ``large'' tables containing 30 or more rows, leading to a decline in accuracy as the table size increases.
While there have been works to decompose both questions and tables using LLMs \citep{10.1145/3539618.3591708}, this line of work still requires providing the full table to the LLM and cannot scale to large tables.
The question studied in this work is if the size of a table can be reduced before passing it to the language model without impacting its performance.


In this work, our aim is to leverage the symbolic representation capabilities of LLMs to reduce table size and their robustness to natural language variations for addressing formatting differences.
Symbolic models, such as text-to-SQL, are not affected by table size and can reliably scale to large tables. However, for reliable storage and querying in a relational database, tables are expected to adhere to a more rigorous formatting.
Tables in the wild, such as those found on the web, often lack this formatting, necessitating substantial preprocessing and normalization efforts to convert the content~\citep{cheng2023binding}. LLMs are well-suited for resolving potential differences in the formating of rows and cell values.
This work aims to strike a balance between table reasoning and table decomposition.
Our approach involves using symbolic methods to narrow down the task to a targeted region in a table and then utilizes LLMs to reason over the limited relevant information.

We propose \textbf{TabSQLify}, a novel approach that integrates symbolic methods with the reasoning power of LLMs. TabSQLify leverages text-to-SQL generation to decompose large tables into smaller and relevant sub-tables for table reasoning tasks. 
The method involves two key steps: \textbf{(1)} generating SQL queries from natural language questions or statements using LLMs under few-shot prompting, then executing the SQL queries on the original tables to obtain sub-tables containing only essential information for answering questions or verifying statements, and \textbf{(2)} using LLMs with the sub-table and the question or claim to generate the answer.
The core concept of the approach is to utilize the natural language understanding and generation strengths of LLMs while reducing their burden in table encoding and reasoning (see Figure \ref{fig:TabSQLify}).
Decomposing tables into sub-tables offers several advantages, including \textbf{(1)} reducing input length for improved scalability and efficiency in reasoning tasks involving large tables, \textbf{(2)} filtering out irrelevant and redundant information that do not contribute to the reasoning process, hence making the reasoning more focused, and \textbf{(3)} providing an intermediate representation (in this case, SQL queries and sub-tables) that is more interpretable and explainable for tracing and verification purposes. 

We evaluate our method on four challenging table reasoning datasets: WikiTQ \citep{pasupat-liang-2015-compositional}, FeTaQA \citep{nan-etal-2022-fetaqa}, TabFact \citep{chen2020tabfact} and WikiSQL \citep{zhongSeq2SQL2017}.
Our evaluation on table-based question answering and fact verification tasks show that our method outperforms other LLM-based baselines, with gpt-3.5-turbo (chatgpt) as the LLM. Moreover, our method can significantly reduce the input length, making it more scalable and efficient for large-scale table reasoning applications than existing methods that require the full table context as input. 

The contributions of this paper are as follows:

\begin{enumerate}
    
    \item We present a novel approach that utilizes text-to-SQL generation to decompose tables into smaller, contextually relevant sub-tables, particularly designed for table reasoning tasks. This method offers a substantial reduction in table size, proving particularly advantageous for large tables that exceed the maximum allowable context window of LLMs. 
    \item Our model outperforms some of the leading models that employ multiple responses and self-consistency. Clearly using those techniques can further boost the performance of our method. 
    \item Our evaluation on challenging table reasoning datasets demonstrates the remarkable performance of our method compared to existing methods that rely on full tables as input. A comprehensive evaluation across various tasks is conducted to elucidate both the advantages and constraints of our approach.
\end{enumerate}

\section{Related Work}
Our work is closely intertwined with the literature on semantic parsing of questions and table schema (also known as text to data) as well as the reasoning applied to semi-structured tables (alternatively known as data to text).

\subsection{Semantic Parsing: Text to Data}
Table-based reasoning conventionally involves semantic parsing of questions and subsequent execution of the generated queries on tables. Traditional models in this domain were often domain-specific, supporting controlled natural language~\citep{popescu2003towards,li2007nalix}, and posed challenges in adaptation to new domains or datasets. However, recent models leveraging machine learning techniques or large language models are trained on extensive datasets and query repositories, supporting a shift towards greater domain-independence. In particular, LLMs, when used with few-shot prompting, serve as powerful code generators, and techniques such as controlled decoding further improves the reliability of code generation~\citep{brown2020language, rajkumar2022evaluating, pourreza2023din, chang2023prompt, ni2023lever}.

Cross-domain benchmarks such as WikiSQL \citep{zhongSeq2SQL2017}, Spider \citep{yu-etal-2018-spider}, CoSQL \citep{yu-etal-2019-cosql}, SParC \citep{yu-etal-2019-sparc}, and BIRD \citep{li2023llm} have played a pivotal role in advancing this field, offering diverse examples of natural language queries paired with formal query language counterparts, such as SQL.

Glass et al. \citep{glass-etal-2021-capturing} innovatively explores methods to capture both row and column semantics, improving the model's query comprehension. Inner Table Retrieval (ITR) \citep{lin-etal-2023-inner} employs a similarity-based approach for locating sub-tables. These approaches involve pre-training and fine-tuning, which heavily rely on specific datasets. This reliance makes them inapplicable without access to a corresponding training dataset, while the need for optimal hyperparameters further limits their generalization.

In this line of work, the reasoning is generally done on questions and table schemata, with the expectation that the data in a table strictly adheres to the table schema (e.g., all values in a column having the same data type).

\subsection{Table Reasoning: Data to Text}

The relevant models can be categorized into more traditional models and recent LLM-based models. Many early models undergo pre-training on both tables and text to acquire a joint representation, utilizing this representation for reasoning without relying on symbolic execution. Notably, TaPas \citep{herzig-etal-2020-tapas} retrieves masked information from tables, TAPEX \citep{liu2022tapex} employs the BART model to emulate an SQL executor, ReasTAP \citep{zhao-etal-2022-reastap} instills reasoning skills via pre-training, TABERT \citep{yin-etal-2020-tabert} encodes a subset of table content most pertinent to the input, and PASTA \citep{gu-etal-2022-pasta} pre-trains language models to be cognizant of common table-based operations. All these models have contributed to the progress on table-based reasoning. Despite achieving commendable results through pre-training on substantial datasets, these models still necessitate fine-tuning on task-specific datasets \citep{chen-2023-large}.

Large language models have become competitive models in many domains and tasks including table reasoning, with their reasoning capabilities covering math, common sense, and symbolic reasoning. This is often done using few-shot prompts without fine-tuning~\citep{brown2020language}. It has been shown that the reasoning capabilities of these models can be further improved by breaking more complex tasks into steps, using methods such as chain-of-thought (CoT) \citep{wei2023chainofthought} and ZeroCoT \citep{kojima2023large} or more carefully selecting examples in the prompt \citep{liu-etal-2022-makes}.
The Table-CoT Model \citep{chen-2023-large} generates the final answer to a question by employing in-context learning and chain-of-thought prompting to table-based tasks. In contrast, the BINDER \citep{cheng2023binding} model generates programs in a programming language, extending its capabilities to solve commonsense problems. The DATER \citep{10.1145/3539618.3591708} approach uses LLMs to decompose tables and questions for solving table-based QA and fact verification tasks.

ReAcTable \citep{zhang2023reactable} adopts the ReAct paradigm, encompassing step-by-step reasoning, code execution through external tools, intermediate table generation, and a majority voting mechanism. This method leverages LLMs to decompose the problem into multiple steps, each consisting of logical operations in the form of code to process tabular data as required. In a more recent model, LEVER \citep{ni2023lever} presents a method to enhance language-to-code generation by training to validate the generated programs based on their execution results. 

StructGPT \citep{jiang-etal-2023-structgpt} enhances LLM reasoning for structured data using an Iterative Reading-then-Reasoning approach. However, the complexity and cost of StructGPT are exacerbated by the practice of passing entire tables to LLM in the reading phase, thus limiting the model's scalability to large tables due to token limits. Chain-of-Table \citep{wang2024chain} extends Chain-of-Thought to tables, improving accuracy by transforming input tables and guiding LLM with intermediate tables during reasoning. However, it requires multiple intermediate steps and LLM calls.

Table reasoning approaches typically operate under the assumption that the tables in question are sufficiently small to be directly input into the model. This specific issue is the focus of our investigation in this paper.




\section{Methodology}

Our approach capitalizes on the proficiency of LLMs in parsing natural language text and generating SQL to enhance their capabilities in table reasoning. Large language models face challenges in accommodating extensive contextual information, especially when dealing with large tables that exceed their token limits. Increasing this size for large tables is unrealistic due to the quadratic time and memory complexities of self-attention mechanism in input length. Furthermore, LLMs are more likely to produce errors when handling lengthy contexts~
\citep{chen-2023-large, 10.1145/3571730}. To overcome these challenges, our work efficiently identifies and extracts relevant table parts, optimizing the input prompt size without sacrificing performance.





\subsection{Table Preprocessing}
\label{sec:preprocessing}
Although tabular data is typically stored in a relational database and queried using SQL, many tables collected from web sources lack the rigorous structure and consistency that is needed for SQL queries to retrieve correct answers. It is generally a challenge to fully clean data from different sources or with no clean lineage records. Our hypothesis is that applying some general table cleaning and relaxing the granularity of retrievals to relevant rows and columns that have the answers, instead of the exact answers, makes the SQL engine more reliable. Of course, the exact answer must be extracted at the end. This is done in our reasoning phase  ($\S$ \ref{sec:method-reasoning}) where an LLM is used, and it is better equipped to handle formatting differences.

For our table cleaning, we normalized  numerical values and date fields. In particular, numerical values frequently feature commas, necessitating preprocessing to ensure consistency. To address this, we uniformly removed commas from all numerical entries. Additionally, the diverse date formats within the tables posed a challenge in generating accurate conditions for SQL queries. To address this, we standardized all date formats to the YYYY-MM-DD format. As an example, we converted numbers like ``360,000'' to ``360000,'' and different date formats such as ``31 October 2008,'' ``31 Oct 2008'' and ``October 31, 2008'' to the standardized ``2008-10-31''.

\subsection{Subtable Selection}
The subtable selection can be done by three strategies: \textbf{(1)} selecting essential columns, \textbf{(2)} selecting essential rows, and \textbf{(3)} selecting both essential columns and rows.
 
In this step, instead of feeding the entire table to the LLM, we provide essential table information such as the title, column names, and three example rows alongside the question. We utilize few-shot learning for this step, where we provide the LLM with a few examples. Subsequently, the LLM generates an SQL query to select the subtable based on this provided information. Selecting essential rows may require performing grouping and aggregation, and our generated SQL queries can include GROUP BY clauses and aggregation functions.

\begin{figure}[!htp]
    \centering
    \resizebox{\columnwidth}{!}{%
    \includegraphics{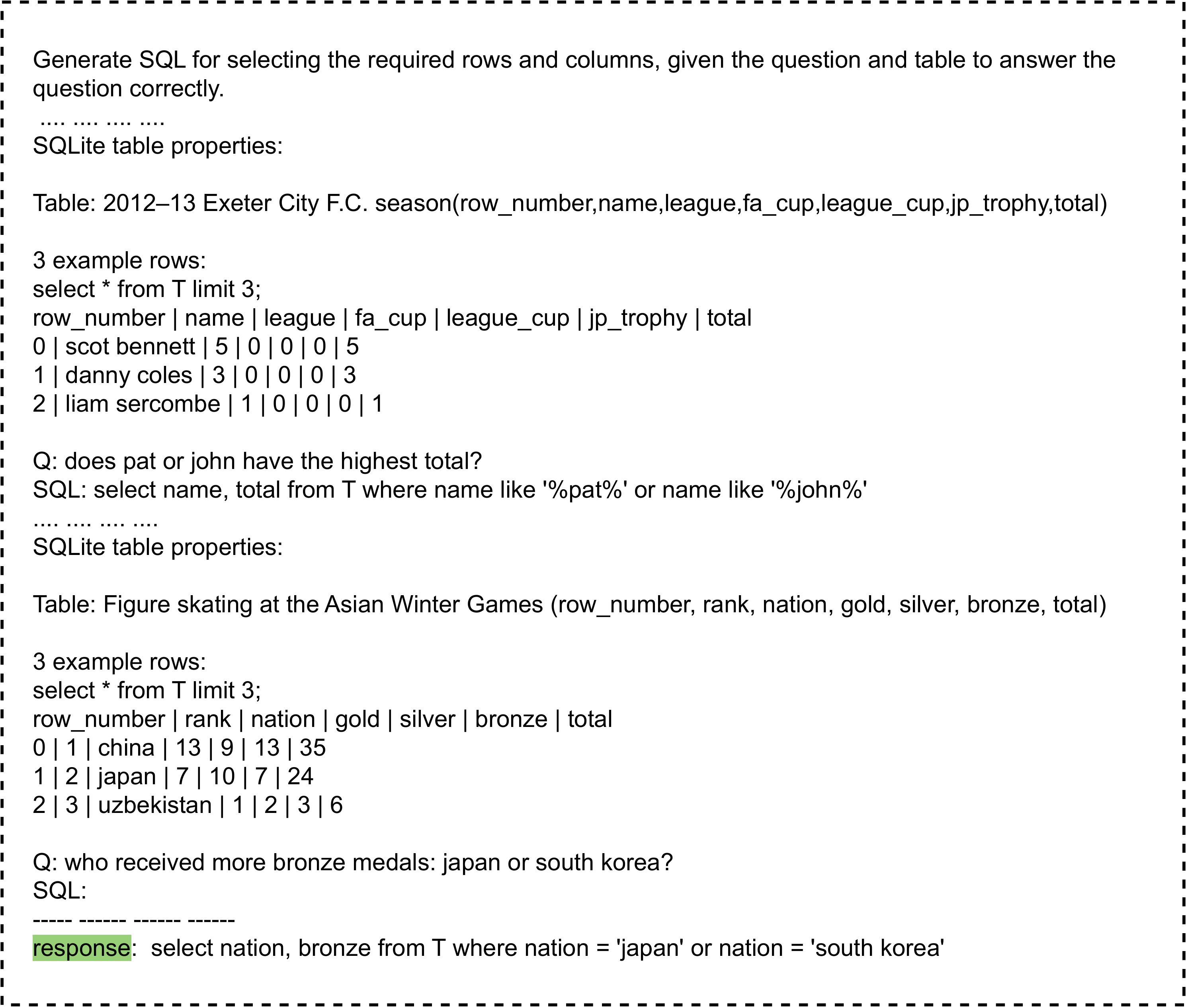}}
    \caption{Prompt used for the subtable selection step of TabSQLify\textsubscript{col+row}.}
    \label{fig:col_row_prompt}
\end{figure}

By selecting the essential columns and rows, we are reducing the context size while optimizing the relevance of information for subsequent reasoning tasks. When employing strategies (2) or (3), sometimes essential rows may not be safely extracted, for example returning an empty table due to noisy input. In those cases, we opt for the column selection strategy. The format of the prompt used for selecting necessary columns and row is described in Figure \ref{fig:col_row_prompt}. 


It is still conceivable that the output of the subtable selection step remains large, for example when finding the top-k most popular products for large values of k. We consider this limitation as inherent to the nature of the task and not specific to our approach since the sub-table containing all the items of the top-k is necessary to answer this question.
 


\subsection{Reasoning and Answer Generation}
\label{sec:method-reasoning}

\begin{figure}[!htp]
    \centering
    \resizebox{\columnwidth}{!}{%
    \includegraphics{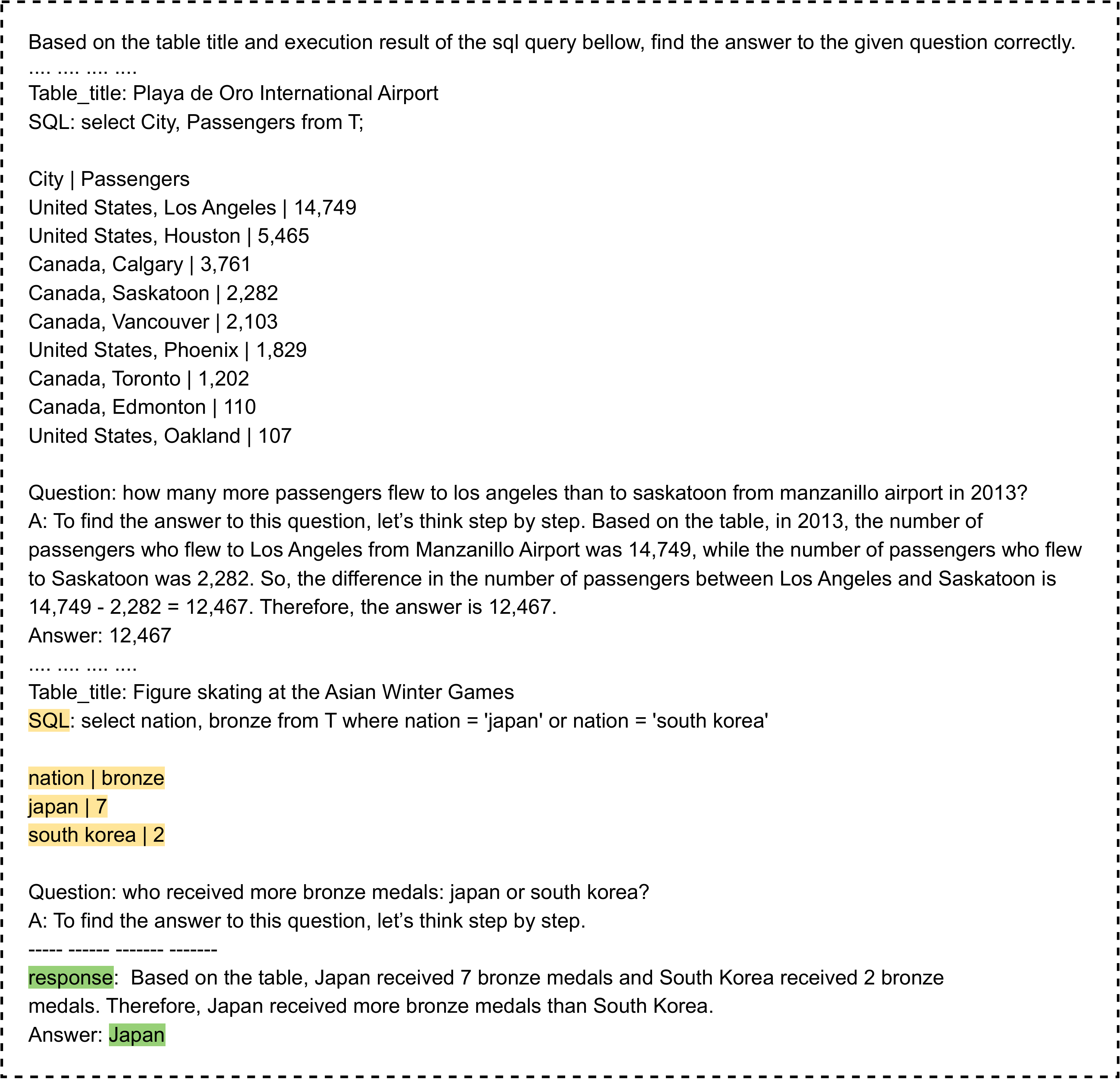}}
    \caption{Prompts used for the answer generation step.}
    \label{fig:ans_promt}
\end{figure}

In this step, an LLM is employed,  wherein we input  the SQL derived from the previous step, the subtable obtained by executing the SQL query and the question. Depending on the domain, additional contextual information, such as the surrounding text, may also be incorporated. This approach is adopted to help the model focus on the relevant parts  for understanding the context and answering the question. Moreover, we utilize few-shot learning techniques while adhering to the Chain-of-Thought prompting style. The format of the answer generation prompt is described in Figure \ref{fig:ans_promt}.

\section{Experimental Setup}

\subsection{Dataset}
We assess our proposed approach across four datasets centered on reasoning with tables. Given our constraints on using LLMs, in terms of the number of requests and associated costs, our method is exclusively evaluated on the test sets of these datasets, with no fine-tuning on the training sets.


\textbf{WikiTQ} WikiTableQuestions (WikiTQ) contains complex questions annotated by crowd workers based on Wikipedia tables. These questions involve multiple complex operations, such as comparison, aggregation, and arithmetic operations, which require reasoning over multiple entries in a table. The standard test set contains 4,344 samples \citep{pasupat-liang-2015-compositional}. 

\textbf{FetaQA} Free-form Table Question Answering (FeTaQA) contains free-form table questions that require deep reasoning and understanding. These questions are usually hard because it requires processing information from different parts of the table. Unlike WikiTQ, this dataset annotates long free-form answers. Our approach is evaluated on the test set that contains 2,003 samples \citep{nan-etal-2022-fetaqa}.

\textbf{TabFact} Table-Fact-Checking (TabFact) is a benchmark for verifying facts based on tables, which includes statements created by crowd workers using tables from Wikipedia. For example, a statement must be judged as either “True” or “False” based on the information in a given table. The accuracy is reported on the test-small set, which contains 2,024 statements and 298 tables \citep{chen2020tabfact}.

\textbf{WikiSQL} WikiSQL is a simpler TableQA dataset, necessitating the filtering and aggregation of information from the table content. Each question in WikiSQL is associated with a ground truth SQL query, from which we extract the gold answer and compare it with our results. We present the accuracy achieved on the test set of WikiSQL \citep{zhongSeq2SQL2017}.



\subsection{Implementation Details}
In the experiments, we use gpt-3.5-turbo (chatgpt) as our language model. The prompt format mainly follows \citet{chang2023prompt} and \citet{tai2023exploring}, which inputs the table schema and the first three table rows. The detail LLM hyper-parameters are provided in Appendix \ref{appendix:hyperparameters}, and all our code and prompts are available at \url{https://github.com/mahadi-nahid/TabSQLify}. 


\subsection{Baselines}
We compare our approach with several strong baseline methods. These methods can be split into two groups. 

\textbf{Pre-training and fine-tuning based models} Our evaluation involves comparing our work with different models ranging from pre-training to fine-tuning. These models, pretrained on a large table corpus, aim to encode a given table as a plain sequence into an encoder and subsequently employ a decoder to generate an answer. As our baselines, we consider Table-BERT \citep{chen2020tabfact}, LogicFactChecker \citep{zhong-etal-2020-logicalfactchecker}, TaPas \citep{herzig-etal-2020-tapas}, SAT \citep{zhang-etal-2020-table}, TAPEX \citep{liu2022tapex}, GraPPa \citep{yu2020grappa}, PASTA \citep{gu-etal-2022-pasta} as our baslines. For FeTaQA evaluation, we compare our results against T5 \citep{raffel2020exploring, nan-etal-2022-fetaqa}.

\textbf{LLM based models} 
For the LLM based methods with in-context learning, we compare against TableCoT \citep{chen-2023-large}, BINDER \citep{cheng2023binding}, DATER \citep{10.1145/3539618.3591708}, StructGPT \citep{jiang-etal-2023-structgpt}, ReAcTable \cite{zhang2023reactable}, ITR \citep{lin-etal-2023-inner}, LEVER \citep{ni2023lever} and Chain-of-Table \citep{wang2024chain} as our baselines.


\subsection{Evaluation metrics}
For the WikiTQ and WikiSQL dataset, exact match (EM) accuracy was used to check if the predicted answers were the same as the correct ones. To account for different formatting of date and number fields, we added a pre-mactching check \cite{cheng2023binding}, consistent with preprocessing ($\S$ \ref{sec:preprocessing}). 
The accuracy of TabFact was determined using binary classification accuracy. To evaluate FeTaQA, metrics such as ROUGE-1, ROUGE-2, and ROUGE-L \citep{lin-2004-rouge} were used. However, ROUGE score lacks the ability to gauge the faithfulness and correctness of model-generated content. In line with \citet{chen-2023-large}, a human evaluation was conducted across four aspects: fluency (assessing linguistic errors), correctness (ensuring accurate answers to questions), faithfulness (verifying grounding on the input table), and adequacy (evaluating the comprehensiveness of the generated sentence in covering all answers) \cite{nan-etal-2022-fetaqa}.


\section{Results}
\subsection{Model accuracy}
As shown on Table \ref{tab:wikitqa}, TabSQLify achieves an accuracy of 62.0\% and 63.7\% on the more challenging WikiTA dataset when reasoning is performed solely using the extracted columns and extracted rows, respectively. By extracting both the necessary columns and rows, we achieve an accuracy of 64.7\%. Our model outperforms all pretrained models and LLM-based baselines, with chatgpt used as the LLM, on WikiTQ dataset~\footnote{Codex was not available at the time of running our experiments, and the reported results are from the respective papers of our baselines.}. It surpasses BINDER-Codex and achieves accuracy very close to the state-of-the-art model DATER. It is worth noting that, unlike our model, which considers only one response, both BINDER and DATER utilize 20 responses for the WikiTQ dataset to obtain the final answer.

\begin{table}[h]
\centering
\small
\begin{tabular}{lc}
\hline
Models                                & Accuracy \\ \hline \hline
\citealp[]{agarwal2019learning}       & 44.1              \\ 
\citealp[]{wang-etal-2019-learning}   & 44.5              \\ 
TaPas                                 & 48.8              \\ 
GraPPa                                & 52.7              \\ 
LEVER                                & 62.9      \\   
ITR                                 & 63.4 \\   \hline 
GPT-3 CoT                            & 45.7              \\ 
TableCoT-Codex                       & 48.8              \\ 
DATER-Codex                                & 65.9            \\ 
BINDER-Codex                         & 61.9        \\
ReAcTable-Codex                      & 65.8           \\ 
SQL-Codex                            & 61.1        \\ \hline 
BINDER-chatgpt                       & 55.4      \\ 
DATER-chatgpt                    & 52.8         \\ 
ReAcTable-chatgpt                    & 52.5         \\
SQL-chatgpt                          & 54.1         \\ 
TableCoT-chatgpt                     & 52.4       \\ 
StructGPT                         & 52.2      \\ 
Chain-of-Table                      & 59.9   \\ 
TabSQLify\textsubscript{col}         & 62.0             \\ 
TabSQLify\textsubscript{row}         & 63.7             \\
TabSQLify\textsubscript{col+row}     & \textbf{64.7}           \\ \hline

\end{tabular}
\caption{Accuracy compared to the baselines on WikiTQ with the official evaluator.}
\label{tab:wikitqa}
\end{table}


For the TabFact dataset, as illustrated in Table \ref{tab:tabfact}, TabSQLify outperforms all LLM-based state-of-the-art approaches, with ChatGPT as the LLM. We achieve an accuracy of 79.5\% when we extract the required sub-table by applying both column and row filtering. It is important to highlight that BINDER and DATER employ multiple responses and self-consistency to obtain the final answer. The reported results on TabFact are based on 50 responses for BINDER, 20 responses for DATER, and only one response for our model, hence it gives a lower bound of our model performance. 

\begin{table}[h]
\centering
\small
\begin{tabular}{lc}
\hline
Model                               & Accuracy \\ \hline \hline
Table-BERT                  & 68.1              \\ 
LogicFactChecker            & 74.3              \\ 
SAT                         & 75.5              \\ 
TaPas                       & 83.9              \\
TAPEX                       & 85.9              \\
SaMoE                       & 86.7              \\ 
PASTA                       & 90.8              \\ 
Human                       & 92.1              \\ \hline 
TableCoT-Codex              & 72.6              \\ 
DATER-Codex                 & 85.6              \\
BINDER-Codex                & 85.1     \\ 
ReAcTable-Codex             & 83.1       \\ \hline
ReAcTable-chatgpt           & 73.1      \\
TableCoT-chatgpt            & 73.1              \\ 
BINDER-chatgpt              & 79.1    \\ 
DATER-chatgpt              & 78.0     \\
Chain-of-Table             & 80.2              \\ 
TabSQLify\textsubscript{col}   & 77.0              \\ 
TabSQLify\textsubscript{row}    & 78.5         \\
TabSQLify\textsubscript{col+row}  & \textbf{79.5}             \\ \hline

\end{tabular}
\caption{Experimental results on TabFact. Here, ``Human'' indicates the human performance \cite{10.1145/3539618.3591708}}
\label{tab:tabfact}
\end{table}
For FeTaQA dataset, we achive a performance comparable to the baselines. As ROUGE metrics do not reflect the actual correctness of the model's responses, we manually evaluated 100 randomly chosen sample and quantified their performance in terms of fluency, correctness, adequacy and faithfulness. The performance is summarized in Tables \ref{tab:fetaqa1} and \ref{tab:fetaqa2}. TabSQLify outperforms models based on fine-tuning and pre-training, such as T5-large. The evaluation suggests that the model's output closely aligns with average human performance in terms of fluency, adequacy, and faithfulness. The correctness is notably impressive, although it falls behind human-level performance. This indicates that, utilizing TabSQLify results in high accuracy without the need for the entire table, showcasing the model's high level of precision in retrieving the relevant sub-table. 
We additionally assess the results using RAGAS \cite{ragas-github}. The evaluation outcomes obtained from RAGAS are detailed in Appendix \ref{appendix:ragas}.

\begin{table}[h]
\centering
\small
\begin{tabular}{lcccc}
\hline
Model                   & R-1           & R-2           & R-L \\ \hline \hline
T5-small                & 0.55          & 0.33          & 0.47  \\ 
T5-base                 & 0.61          & 0.39          & 0.51   \\ 
T5-large                & 0.63          & 0.41          & 0.53  \\ \hline 
TableCoT-Codex          & 0.62          & 0.40          & 0.52   \\ 
DATER-Codex                   & 0.66          & 0.45          & 0.56   \\ 
ReAcTable               & 0.71          & 0.46          & 0.61 \\ 
TableCoT-chatgpt        & 0.62          & 0.39          & 0.51    \\ 
TabSQLify\textsubscript{col}           & 0.57          & 0.34          & 0.47  \\ 
TabSQLify\textsubscript{row}           & 0.60          & 0.37       & 0.49  \\ 
TabSQLify\textsubscript{col+row}       & 0.58          & 0.35          & 0.48   \\ \hline
 
\end{tabular}
\caption{Experimental results on FeTaQA.}
\label{tab:fetaqa1}
\end{table}


\begin{table}[H]
\centering
\small
\scalebox{0.75}{
\begin{tabular}{lcccc}
\hline
Model                    &Fluency       &Correct        &Adequate       &Faithful \\ \hline \hline
T5-large                 & 94.6         & 54.8          & 50.4          & 50.4     \\ 
Human \cite{chen-2023-large} & 95       & 92.4          & 95.6           & 95.6 \\ 
TableCoT-chatgpt              & 96         & 82         & 75            & 87     \\ \hline
TabSQLify\textsubscript{col}   & 98         & 83         & 79           & 85    \\ 
TabSQLify\textsubscript{row}     & 96         & 80          & 77          & 89    \\ 
TabSQLify\textsubscript{col+row}   & 97        & 88          & 84          & 93    \\ \hline

\end{tabular}
}
\caption{Human evaluation results on FeTaQA.}
\label{tab:fetaqa2}
\end{table}

\begin{table}[]
\centering
\small
\begin{tabular}{lc}
\hline
Model                       & Accuracy \\ \hline \hline
SEQ2SQL                      & 59.4\% \\ 
StructGPT                   & 65.6\%  \\ 
RCI \cite{glass-etal-2021-capturing}  & 89.8\%  \\ 
TabSQLify\textsubscript{col+row}  & \textbf{76.7\%} \\ \hline
\end{tabular}

\caption{Experimental results on WikiSQL. RCI is a fine tuning based model, and its results may not be directly comparable due to the model’s high reliance on the training set.}
\label{tab:wikisql}
\end{table}

TabSQLify shows an outstanding performance on the WikiSQL dataset, as demonstrated in Table \ref{tab:wikisql}. This dataset appears to be easier compared to the WikiTQ test dataset, with our approach achieving 76.7\% accuracy. In 70\% of cases, it can produce the answer in the first step, eliminating the need to pass the sub-table and question for the second step.


\subsection{Scalability and robustness }
\label{sec:scalability-experiment}
We assessed the scalability and robustness of our model by imposing a token limit on each table across three datasets: WikiTQ, FeTaQA and TabFact. To accomplish this, we established cutoff thresholds to discard tokens exceeding these limits. Subsequently, we evaluated the model's performance within these constrained token boundaries. For the WikiTQ dataset, we set the cutoff threshold at 2000, while for both the TabFact and FeTaQA datasets, it was set to 600. Table \ref{tab:limit_tok_sample} summarizes the distribution across different classes, illustrating the categories based on the percentage of discarded table tokens (see Appendix \ref{appendix:scalability} for more detail).

\begin{table}[H]
\centering
\small
\begin{tabular}{lcccc}
\hline
Cut-off (\%)              & WikiTQ       & FeTaQA      &TabFact\\ \hline \hline
0 - 10\%                  & 76             & 81           & 91      \\ 
10 - 25\%                 & 89             & 143          & 141       \\ 
25 - 50\%                 & 116            & 202          & 260         \\ 
50\% +                    & 128            & 69           & 81        \\\hline
\end{tabular}
\caption{The distribution of samples across various classes as a function of the percentage cut-off of table tokens.}
\label{tab:limit_tok_sample}
\end{table}

The evaluation results for the WikiTQ dataset are presented in Table \ref{tab:limit_tok_result_wtq}. Our model consistently performs well within the specified token boundary. In contrast, the performance of TableCoT is subpar. We have observed a similar trend in the other two datasets (see Tables \ref{tab:limit_tok_result_tf} and \ref{tab:limit_tok_result_fqa}).

\begin{table}[h]
\centering
\small
\begin{tabular}{lcccc}
\hline
Cut-off (\%)             & TableCoT            &TabSQLify\textsubscript{col+row}\\ \hline \hline
0 - 10\%                  & 40.7             & 64.4  \\ 
10 - 25\%                 & 49.4             & 60.6          \\ 
25 - 50\%                 & 46.5             & 66.3                  \\ 
50\% +                   & 33.3             & 56.2                  \\\hline
\end{tabular}
\caption{Performance across different classes based on the percentage cut-off of table tokens in the WikiTQ dataset.}
\label{tab:limit_tok_result_wtq}
\end{table}

\begin{table}[h]
\centering
\small
\begin{tabular}{lcccc}
\hline
Cut-off (\%)         & TableCoT            &TabSQLify\textsubscript{col+row}\\ \hline \hline
0 - 10\%               & 76.9                 & 79.1     \\ 
10 - 25\%              & 67.3                 & 80.8           \\ 
25 - 50\%              & 63.0                 & 70.0              \\ 
50\% +                & 55.5                 & 72.8               \\\hline
\end{tabular}
\caption{Performance across different classes based on the percentage cut-off of table tokens in the TabFact dataset}
\label{tab:limit_tok_result_tf}
\end{table}

\begin{table}[H]
\centering
\small
\begin{tabular}{llll}
\hline 
\multicolumn{4}{c}{TableCoT}   \\ \hline \hline
Cut-off (\%)    & R-1    & R-2      & R-L \\ \hline
0 -10\%        & 0.58    & 0.35    & 0.45   \\
10-25\%        & 0.60     & 0.37   & 0.50  \\
25-50\%        & 0.53   & 0.30     & 0.43   \\
50\% +         & 0.49   & 0.28    &  0.40   \\ \hline
\multicolumn{4}{c}{TabSQLify\textsubscript{col+row}}  \\ \hline \hline
Cut-off (\%)    & R-1    & R-2      & R-L \\ \hline
0 -10\%        & 0.62   & 0.39    & 0.50   \\
10-25\%        & 0.64   & 0.42   &  0.53  \\
25-50\%        & 0.55   & 0.32   & 0.44  \\
50\% +         & 0.51   & 0.31    & 0.41   \\ \hline
\end{tabular}
\caption{Performance across different classes based on the percentage cut-off of table tokens in the FeTaQA dataset}
\label{tab:limit_tok_result_fqa}
\end{table}

In the WikiTQ dataset, 128 tables contain >4000 tokens exceeding chatGPT's maximum token limit (4096 tokens including table and question). Table \ref{tab:wiki-large} reports the performance on these instances. These results reveal that both BINDER \cite{cheng2023binding} and DATER \cite{10.1145/3539618.3591708} face challenges when dealing with large tables. Specifically, BINDER-Codex achieves only 29.6\% accuracy, while DATER achieves an accuracy of 34.6\%. BINDER-chatgpt fails to produce any correct answers for these large tables. On the other hand, Chain-of-Table \cite{wang2024chain} achieves an accuracy of 44.8\%.

In contrast, our model outperforms these baselines significantly. It is crucial to note that our Table-CoT achieves this accuracy because the answers for questions about those large tables are typically in the upper part, fitting within the LLM's context boundary. If the answer is elsewhere, all models fail. On the other hand, our model has no issue with the answer's position in a table, making it scalable for large tables. 

\begin{table}[h]
\centering
\small
\begin{tabular}{lc}
\hline
Model                       & Acc (Large)\\ \hline \hline
BINDER-Codex                & 29.6     \\ 
BINDER-chatgpt             & 0.0      \\ 
DATER-chatgpt              & 34.6 \\ 
Table-CoT-chatgpt           & 35.1      \\
Chain-of-Table \cite{wang2024chain}   & 44.8  \\ 
TabSQLify\textsubscript{col}              & 50.0     \\ 
TabSQLify\textsubscript{row}              & \textbf{57.0}      \\
TabSQLify\textsubscript{col+row}           & 52.3      \\ \hline

\end{tabular}
\caption{Experimental results on Large (>4000 tokens) tables from WikiTQ. As the input tables grow larger, we observe a decline in performance for strong baseline models.}
\label{tab:wiki-large}
\end{table}

\begin{figure}[H]
    \centering
    \resizebox{\columnwidth}{!}{%
    \includegraphics{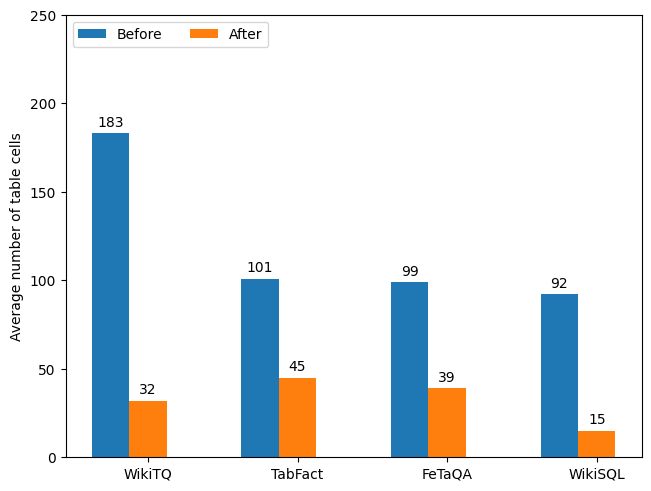}}
    \caption{Reduction in table size using our row-col filtering across four datasets, showing a significant reduction of the table size.}
    \label{fig:reduction}
\end{figure}



\subsection{Table size reduction}

Figure \ref{fig:reduction} demonstrates the average reduction in the number of table cells before and after employing TabSQLify\textsubscript{col+row} across three datasets. This reduction, from 183 to 32 cells in WikiTQ, indicates a substantial decrease in sub-table size while maintaining a strong performance. Likewise, similar trends can be observed in the TabFact, FetaQA and WikiSQL datasets. When utilizing both column and row filters to extract the required subtable, direct answers to questions may be found. Specifically, in the WikiTQ dataset, TabSQLify\textsubscript{col+row} successfully retrieves answers in 58\% of cases by executing the generated query, requiring the answer generation step only for the remaining 42\% of cases. 


\subsection{Error Analysis}

An important advantage of TabSQLify is its ability to provide the intermediate stages of reasoning path, including SQL queries and sub-tables. To conduct our error analysis, we randomly selected 100 responses generated by TabSQLify\textsubscript{row+col} from the WikiTQ and TabFact test sets. The identified errors are categorized into incorrect columns, incorrect conditions, incorrect reasoning, and false negatives, as listed in Table \ref{tab:error_wtq}. 

In this context, a ``missing column'' refers to instances where TabSQLify either selects incorrect columns or omits necessary columns to answer the question. ``missing rows'' denotes situations where the generated SQL query contains an erroneous condition within the WHERE clause. Cases where the extracted sub-table is adequate to answer the question, but the LLM fails to provide a correct response, are labeled as ``incorrect reasoning''. Additionally, within the dataset, there are instances where the gold answer is incorrect or misjudged by the evaluator, which we classify as ``incorrect annotation''. 

From the table, we observe that out of 100 error cases from WikiTQ, 6\% involve the generated SQL query missing columns, while 56\% miss required rows. The irregular format of the text in the table is identified as the primary cause. Additionally, in 29\% of cases, the reasoning is found to be incorrect, while 9\% exhibit incorrect annotation. In the TabFact\ dataset, 10\% of the time, the subtable selection query misses required columns, and in 32\% of cases, it misses required rows. The main source of errors is incorrect reasoning, accounting for 50\% of cases, while 8\% involve incorrect annotations.

\begin{table}[h]
\centering
\small
\begin{tabular}{lcc}
\hline
Error Type               & WikiTQ   & TabFact   \\ \hline \hline
Missing Columns          & 6\%           & 10\%     \\
Missing Rows              & 56\%         & 32\%    \\ 
Incorrect Reasoning       & 29\%         & 50\%     \\
Incorrect Annotation      &  9\%         & 8\%  \\ \hline
\end{tabular}
\caption{Error types of 100 samples from WikiTQ and TabFact of TabSQLify\textsubscript{col+row}}
\label{tab:error_wtq}
\end{table}

\section{Conclusion}

Our proposed decomposition approach has shown promise across different table reasoning tasks, achieving remarkable performance compared to models that require the use of a full table. Our method is novel in leveraging text-to-SQL generation to decompose tables into smaller and relevant sub-tables tailored for table reasoning tasks. This approach provides a new perspective and direction for table reasoning research, and we hope it will inspire more future work on combining natural language understanding and structured data processing. 

\section*{Limitations}
Our approach is not without its limitations. While it shows promise in reducing table size and maintaining a strong performance, for large tables, the size of  a column can exceed the context window size, and the approach may not be applicable. Also, after our preprocessing, the tables are stored in a relational tables. For less regular tables, more preprocessing may be needed.

\section*{Acknowledgements}
We extend our sincere gratitude to all anonymous reviewers for their invaluable feedback, insightful suggestions, and positive remarks about our work. This research has been supported by the Natural Sciences and Engineering Research Council of Canada. Also, Md Mahadi Hasan Nahid was supported by the Alberta Innovates Graduate Student Scholarship.

\section*{Ethical Considerations}
The datasets utilized in this study are accessible through peer-reviewed articles, as specified in the references. Our source code is made openly available for future research under the MIT License. It's important to note that since our framework relies on gpt-3.5-turbo, it may inherit ethical concerns associated with gpt models, such as potential responses to toxic content or displaying a biased behavior. 

\bibliography{anthology,custom}
\bibliographystyle{acl_natbib}

\appendix

\section{LLM Hyper-parameters}
\label{appendix:hyperparameters}
We configured the in-context learning hyperparameters for gpt-3.5-turbo according to the specifications outlined in Table \ref{tab:hyper-parameters1} and Table \ref{tab:hyper-parameters2}.

\begin{table}[H]
\centering
\small
\begin{tabular}{lcccc}
\hline
\multicolumn{4}{c}{Sub table selection using Text-to-SQL} \\
Parameter             & WTQA           & FeTaQA        & TabFact \\ \hline \hline
temperature           & 0.3             & 0.3           & 0.3       \\ 
top\_p                & 1               & 1             & 1          \\
sample\_n             & 1               & 1             & 1    \\ 
max\_tokens           & 100             & 100           & 100    \\ 
num\_shots            & 10              & 6            & 8    \\ \hline

\end{tabular}
\caption{Our hyper-parameter setting of LLM for selecting required column/row}
\label{tab:hyper-parameters1}
\end{table}

\begin{table}[H]
\centering
\small
\begin{tabular}{lcccc}
\hline
\multicolumn{4}{c}{Answer Generation} \\
Parameter            & WTQA      & FeTaQA    & TabFact \\ \hline \hline
temperature          & 0.7          & 0.7       & 0.6     \\ 
top\_p               & 1            & 1         & 1      \\   
sample\_n            & 1            & 1         & 1    \\ 
max\_tokens          & 200          & 64        & 100    \\ 
num\_shots           & 2            & 6         & 4    \\ \hline

\end{tabular}
\caption{Our hyper-parameters setting of LLM for the answer generation}
\label{tab:hyper-parameters2}
\end{table}

\section{Scalability and Robustness Experiment}
\label{appendix:scalability}

LLMs function within a restricted token boundary, allowing us to provide only a limited number of tokens as a prompt to the LLM. We report the impact of the cutoff threshold where tokens beyond the cutoff points are discarded ($\S$ \ref{sec:scalability-experiment}).

The cutoff percentage denotes the percentage of tokens that are truncated when the threshold is applied. For example, if a table has 4500 tokens and we set the threshold at 2000, then 2500 tokens of the original table are truncated, and the percentage is 2500/4500 = 55.56\%. We separated the number of samples in different cutoff ranges (see table 5) and compared the results of those samples from different cutoff ranges in table 6 and 7.

For the WikiTQ dataset, we set the threshold at 2000 tokens. In this case, out of 4,344 samples, there are 128 samples where more than 50\% of the tokens of the original table are truncated if we want to pass the original table to the LLM with a maximum token boundary of 2000. In our approach, TabSQLify selects the relevant limited subtable from the original table for a given question. This allows us to fit the subtable within the maximum token boundary when passing it to the LLM, resulting in improved performance. The aim of this experiment is to demonstrate that TabSQLify can be useful under limited token (context) boundary conditions.



\section{Comparison with other models}
\label{appendix:comparison}
In this section, we conduct a comparative analysis of our model against two strong baselines, DATER \cite{10.1145/3539618.3591708} and BINDER \cite{cheng2023binding}.
DATER utilizes Large Language Models (LLMs) for decomposing both questions and tables. On the other hand, BINDER stands out by offering an Application Programming Interface (API) that extends language model (LM) functionalities to programming languages such as SQL and Python. This extension broadens its grammar coverage, enabling the model to address a more diverse range of questions. However, a drawback is that both DATER and BINDER necessitates sending the entire table to the LLM and face challenges when dealing with large tables. Both DATER and BINDER leverage self-consistency \cite{wang2023selfconsistency} strategies to bolster their performance, ensuring a higher level of consistency in their responses.



In our experiment we did not consider using self-consistence decoding strategy. Using self-consistency we can push the performance even higher. Our implementation does not require any additional processing on the SQL code, unlike BINDER, which necessitates a complex re-implementation of the SQL executor \cite{ni2023lever}. BINDER generates a total of 50 samples for a given table and question in the intermediate stages (Generate Neural-SQL: 50); while DATER generates 100 samples in its intermediate stages (table decomposition: 40; Generate Cloze: 20; Generate SQL: 20; reasoning: 20). In contrast, TabSQLify generates only two samples in total, making it simpler and more cost-effective. We summarize the comparison with DATER and BINDER in Table \ref{tab:model_comparison}. 

\begin{table}[h]
\centering
\small
\begin{tabular}{lccc}
\hline
-                   & DATER    & BINDER     & TabSQLify \\ \hline \hline
\# stage            & 4        & 2          & 2           \\ 
Max context size    & 8000     & 8000        & 4096     \\ 
\# of generated samples & 100    & 50    & 2 \\
sampling\_n         & 20-50      & 20       & 1     \\ 
Self Consistency    & yes      & yes         & no         \\ 
Table required      & full     & full           & partial     \\
Cost                & high     & high           & low    \\ \hline
\end{tabular}
\caption{Comparison with the other LLM based models. TabSQLify is much simpler than the other approach.}
\label{tab:model_comparison}
\end{table}

Compared to the other LLM-based approach, our approach has several benefits: (1) Unlike other models our approach do not need to provide the table data to LLM to select the target portion of the table. Instead we utilize text-to-sql capability of LLMs. (2) Our approach requires partial table, not full table. (3) Our model can be applied in tight token boundary (4) Considering only one response our model can achive comparable performance while other top performing model uses more than 20 responses (5) Our approach is less costly and it requires less LLM calls which can be vital factor to reduce the cost. 

\section{RAGAS Evaluation for FeTaQA}
\label{appendix:ragas}
Apart from human evaluation, we analyze 100 sample outputs from the FeTaQA dataset using the RAGAS evaluator \cite{ragas-github}, a framework specifically designed for evaluating Retrieval Augmented Generation (RAG) pipelines. The RAGAS evaluation results is listed in Table \ref{tab:ragas_eval}. RAGAS assesses several key aspects: (1) Faithfulness: Evaluates the factual consistency of the answer concerning the context based on the question, (2) Context Precision: Measures the relevance of the retrieved context to the question, reflecting the quality of the retrieval pipeline, (3) Answer Relevancy: Assesses the relevance of the answer to the question, and (4) Context Recall: Measures the retriever's capability to retrieve all essential information required to answer the question. The performance of TabSQLify is comparable to that of Table-CoT-chatgpt, which utilized the full table context. Additionally, the RAGAS evaluation shows a similar trend to our human evaluation. 

\begin{table}[H]
\centering
\small
\scalebox{0.7}{
\begin{tabular}{lcccc}
\hline
Model                      &Precision    &Recall    &Relevancy  &Faithfulness \\ \hline \hline
TableCoT-chatgpt                 & 0.44        & 0.94          & 0.94          & 0.73     \\ 
TabSQLify\textsubscript{col}     & 0.42         & 0.92         & 0.93          & 0.67    \\ 
TabSQLify\textsubscript{row}      & 0.45        & 0.97        & 0.94          & 0.73   \\ 
TabSQLify\textsubscript{col+row}  & 0.44        & 0.94         & 0.94         & 0.72    \\ \hline
\end{tabular}
}
\caption{RAGAS evaluation results on FeTaQA.}
\label{tab:ragas_eval}
\end{table}

\end{document}